\documentclass[10pt,twocolumn,letterpaper]{article}

\usepackage[pagenumbers]{cvpr} 

\usepackage{graphicx}
\usepackage{amsmath}
\usepackage{mathtools}
\usepackage{iac_pkg}
\usepackage{lipsum}
\usepackage{amssymb}
\usepackage{pifont}
\usepackage{multirow}
\usepackage{tabularx, booktabs}
\usepackage{color}
\usepackage{xspace}
\usepackage{overpic}
\usepackage{arydshln}
\usepackage{subcaption}
\usepackage{xcolor}
\usepackage{bm}

\usepackage{cuted}

\definecolor{citecolor}{RGB}{34,139,34}
\definecolor{lightred}{RGB}{255,100,100}
\definecolor{cell_bisque}{rgb}{1.0, 0.89, 0.77}
\definecolor{cell_blond}{rgb}{0.98, 0.94, 0.75}
\definecolor{cell_blue}{RGB}{155, 187, 228}
\definecolor{princetonorange}{rgb}{1.0, 0.56, 0.0}
\definecolor{pinkpearl}{rgb}{0.91, 0.67, 0.81}
\definecolor{mossgreen}{rgb}{0.68, 0.87, 0.68}

\newcommand{\Paragraph}[1]{\vspace{-0mm}\noindent\textbf{#1.}\hspace{0mm}}
\newcommand{\Section}[1]{\vspace{-1mm} \section{#1} \vspace{-0mm}}

\usepackage{hyperref}

\hypersetup{
    breaklinks=true,
    colorlinks=true,
    citecolor=cyan
}
\usepackage[capitalize]{cleveref}

\def\paperID{123} 
\def\confName{CVPR}
\def\confYear{2026}

\begin{document}
\title{HistoFusionNet: Histogram-Guided Fusion and Frequency-Adaptive Refinement for Nighttime Image Dehazing}


\author{
Mohammad~Heydari\footnotemark[1]\hspace{12pt}
Wei~Dong\footnotemark[1]\hspace{12pt}
Shahram~Shirani\hspace{12pt}
Jun~Chen\hspace{12pt}
Han~Zhou\footnotemark[2]\\
\normalsize{McMaster University}\\
\texttt{\small \{heydam3, dongw22, shirani, chenjun, zhouh115\}@mcmaster.ca}
}


\twocolumn[{
\maketitle
\vspace{-8mm}
\begin{center}
    \setlength{\abovecaptionskip}{2mm}
    \includegraphics[width=\textwidth]{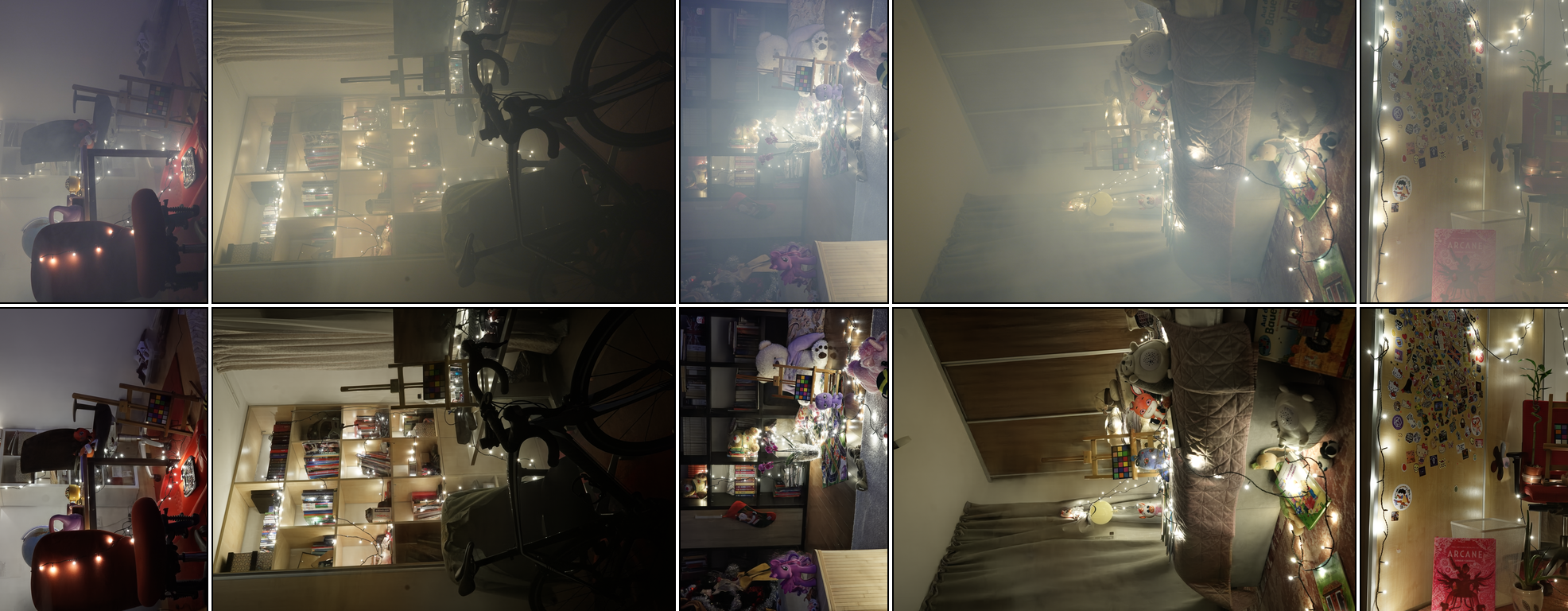}

    \captionof{figure}{Test results of our method on the NTIRE 2026 Nighttime Image Dehazing Challenge~\cite{ntire2026dehazing}. Our \textbf{HistoFusionNet} achieves the \textbf{best} performance among 22 participating teams and generates visually compelling outputs with faithful colors and enhanced structural details.}
    \label{fig:figure_challenge}
\end{center}
\vspace{2mm}
}]

\begingroup
\renewcommand\thefootnote{\fnsymbol{footnote}}
\footnotetext[1]{Equal contribution.}
\footnotetext[2]{Corresponding author.}
\endgroup

\begin{abstract}
\vspace{-2mm}
Nighttime image dehazing remains a challenging low-level vision problem due to the joint presence of haze, glow, non-uniform illumination, color distortion, and sensor noise, which often invalidate assumptions commonly used in daytime dehazing. To address these challenges, we propose \textbf{HistoFusionNet}, a transformer-enhanced architecture tailored for nighttime image dehazing by combining histogram-guided representation learning with frequency-adaptive feature refinement. Built upon a multi-scale encoder--decoder backbone, our method introduces histogram transformer blocks that model long-range dependencies by grouping features according to their dynamic-range characteristics, enabling more effective aggregation of similarly degraded regions under complex nighttime lighting. To further improve restoration fidelity, we incorporate a frequency-aware refinement branch that adaptively exploits complementary low- and high-frequency cues, helping recover scene structures, suppress artifacts, and enhance local details. This design yields a unified framework that is particularly well suited to the heterogeneous degradations encountered in real nighttime hazy scenes. Extensive experiments and highly competitive performance of our method on the NTIRE 2026 Nighttime Image Dehazing Challenge benchmark demonstrate the effectiveness of the proposed method. Our team ranked \textbf{1st} among 22 participating teams, highlighting the robustness and competitive performance of HistoFusionNet. The code is available at \url{https://github.com/heydarimo/Night-Time-Dehazing}. 
\end{abstract}

\section{Introduction}
\vspace{-1mm}
\label{sec:intro}

Nighttime image dehazing aims to recover clear and visually faithful images from haze-degraded scenes under low-light conditions. Compared with daytime dehazing, nighttime restoration is more challenging due to the joint presence of haze scattering, glow around artificial light sources, severe non-uniform illumination, color distortion, and sensor noise. These degradations reduce visibility and perceptual quality, while also impairing downstream tasks such as detection, tracking, and scene understanding in safety-critical applications including autonomous driving, surveillance, and intelligent transportation systems. As a result, robust nighttime dehazing remains an important yet still insufficiently solved problem in image restoration~\cite{cong2024sfsnid,nighthaze,li2015nighttime,liu2022vdm}.

Early dehazing methods were largely built on the atmospheric scattering model (ASM), which relates hazy image formation to scene radiance, transmission, and atmospheric light. While this formulation inspired many prior-based methods, its assumptions are often violated in nighttime scenes, where multiple colored light sources, spatially varying illumination, and glow produce highly heterogeneous degradations. As a result, classical model-based approaches often struggle to generalize to nighttime dehazing, especially when haze and illumination are both complex and locally varying~\cite{ASM,fattal2008dehazing,he2009dcp,berman2016nonlocal,guo2022dehamer,li2015nighttime}.

Recent advances in deep learning have significantly improved image dehazing by learning direct mappings from degraded images to clean targets. CNN-based methods such as DehazeNet, AOD-Net, Gated Fusion Network, FFA-Net, and DCPDN showed that learned representations can outperform hand-crafted priors, especially under complex haze distributions~\cite{cai2016dehazenet,aodnet,gfn,ffanet,dcpdn}. Subsequent works further improved restoration through more effective architectures, including AECR-Net and Trident Dehazing Network, as well as stronger designs for challenging benchmarks~\cite{aecrnet,tdn,griddehazenet,msbdn,psd,benchmark_reside}. However, most of these methods were developed for daytime or general dehazing, and remain limited for the coupled degradations of nighttime scenes.

Transformer-based dehazing methods have recently gained attention for their ability to model long-range dependencies and global context. DeHamer introduces transmission-aware 3D position embedding for dehazing~\cite{guo2022dehamer}, while DehazeFormer and Restormer demonstrate the effectiveness of hierarchical transformer designs for image restoration~\cite{dehazeformer,restormer}. Challenge-oriented methods further explore stronger hybrid backbones. For example, DWT-FFC combines wavelet decomposition, Fourier convolution, and ConvNeXt priors for non-homogeneous dehazing~\cite{dwtffc}, whereas DehazeDCT uses deformable convolution based transformer-like blocks for adaptive aggregation and improved efficiency~\cite{dehazedct}. Despite these advances, nighttime restoration remains challenging because global context alone is often insufficient when degradations are closely tied to dynamic-range variation, local illumination imbalance, and frequency-specific corruption~\cite{aln,sfhformer,frdiff}.

\begin{figure*}[!t]
    \setlength{\abovecaptionskip}{2mm}
    \centering
    \begin{overpic}[width=0.9\textwidth]{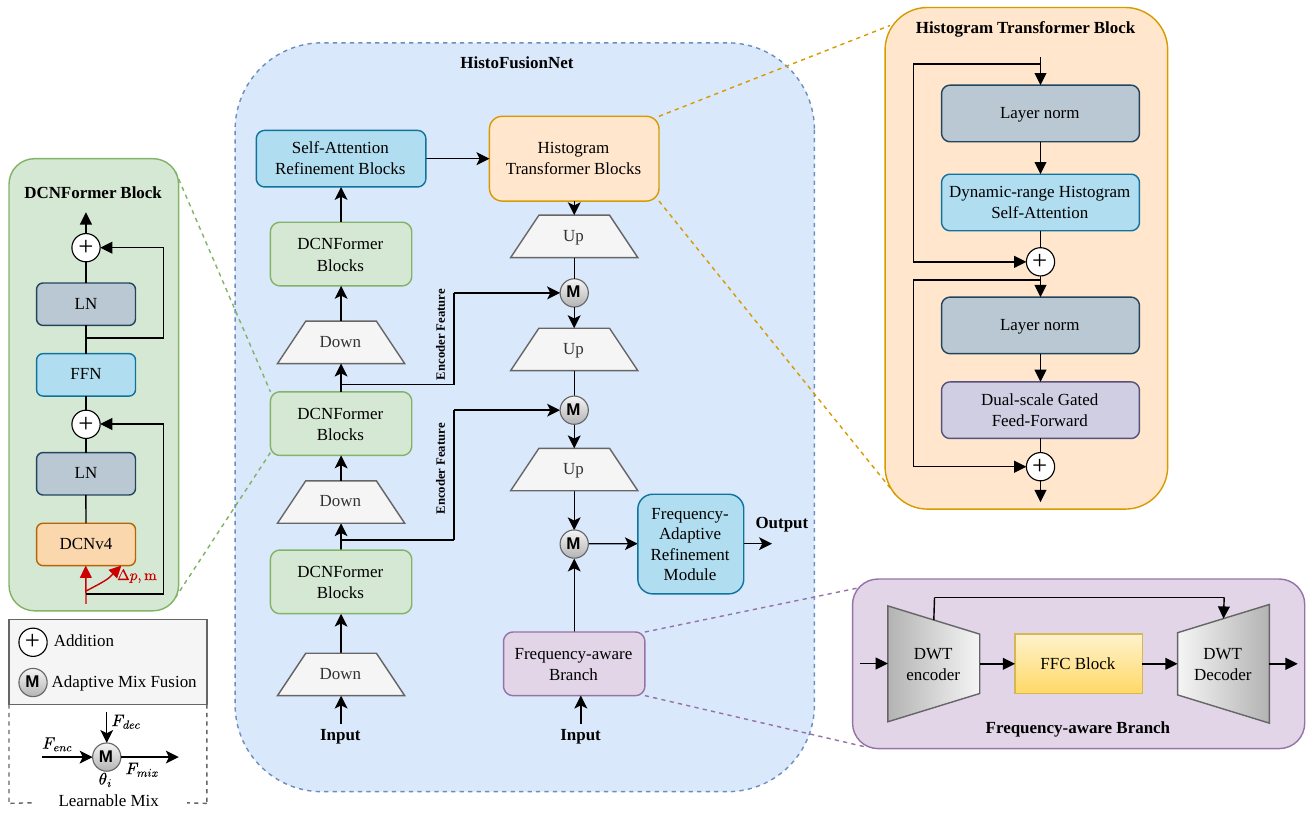}
    \end{overpic}
    \caption{ Overall architecture of \textbf{HistoFusionNet}. Our dehazing network adopts a U-shaped design with a DCNv4-based main branch and an auxiliary frequency-aware branch. Histogram transformer blocks are inserted at the bottleneck to perform dynamic-range aware global aggregation, while a lightweight frequency-adaptive refinement module is employed to enhance color fidelity and recover fine details.} 
    \vspace{-2mm}
    \label{fig_frame}
\end{figure*}

Nighttime-specific methods have begun to address this problem more directly. SFSNiD introduces a semi-supervised nighttime dehazing framework with spatial-frequency interaction and realistic brightness constraints, highlighting the need to jointly handle haze, glow, and noise~\cite{cong2024sfsnid}. NightHaze further shows that self-prior learning can improve robustness on real nighttime degradations~\cite{nighthaze}. Earlier works also emphasized the importance of explicitly modeling glow and illumination effects~\cite{li2015nighttime,liu2022vdm}. These studies suggest that effective nighttime dehazing should account for both degradation-aware brightness variation (similar to low-light image enhancement methods~\cite{zhou2025lita}) and complementary frequency information. However, there remains room for architectures that more directly exploit the structural regularities induced by haze and illumination while maintaining strong restoration fidelity.

Motivated by these observations, we propose \textbf{HistoFusionNet}, a nighttime image dehazing network that combines histogram-guided feature modeling with frequency-adaptive refinement in a unified encoder--decoder framework. Our design is based on two key insights. Adverse-weather degradations often induce similar dynamic-range distortions across spatially distant regions, and nighttime haze affects low- and high-frequency components differently. Accordingly, HistoFusionNet integrates DCNFormer-based multi-scale feature extraction, histogram transformer blocks for dynamic-range aware representation learning, and a frequency-aware refinement module for detail recovery and visual enhancement. Histogram self-attention has proven effective for adverse-weather restoration~\cite{histoformer}, while frequency mining and modulation improve adaptation across degradations~\cite{adair}. Our method combines these ideas in a dehazing architecture tailored to nighttime scenes.

Our main contributions are summarized as follows:

$\diamond$ We propose \textbf{HistoFusionNet}, an effective nighttime image dehazing architecture built on Deformable Convolution v4 (DCNv4)~\cite{xiong2024efficient}, together with histogram-guided fusion and frequency-adaptive refinement within a unified framework.

$\diamond$ We introduce histogram transformer blocks for nighttime image dehazing to capture long-range dependencies among regions with similar dynamic-range degradation, enabling more effective restoration under non-uniform illumination and complex haze.

$\diamond$ We incorporate a frequency-adaptive refinement module to exploit complementary low- and high-frequency information, improving structural recovery, color fidelity, and fine detail restoration.

$\diamond$ Extensive experiments verify the effectiveness of our method. Comparisons with four recent baselines and evaluations on multiple real-world hazy datasets demonstrate the strong performance and generalization capability of \textbf{HistoFusionNet}.

\section{Related Work}
\label{sec:related}

\Paragraph{Nighttime Image Dehazing}
Single image dehazing has been widely studied from both model-based and learning-based perspectives. Early deep methods such as DehazeNet~\cite{cai2016dehazenet}, AOD-Net~\cite{aodnet}, Gated Fusion Network~\cite{gfn}, FFA-Net~\cite{ffanet}, and DCPDN~\cite{dcpdn} demonstrated the effectiveness of end-to-end feature learning over hand-crafted priors. Subsequent methods further improved restoration quality through stronger architectures, including AECR-Net~\cite{aecrnet}, Trident Dehazing Network~\cite{tdn}, and MSBDN~\cite{msbdn}. More recently, research has shifted toward more challenging real-world and non-homogeneous settings, as reflected by the NTIRE dehazing benchmarks and challenge reports~\cite{ntire2020dehazing,ntire2021dehazing,ntire2023dehazing,ntire2024dnh}. Compared with daytime dehazing, nighttime dehazing is more difficult due to the joint presence of haze, glow, non-uniform illumination, color distortion, and sensor noise. To address these issues, SFSNiD~\cite{cong2024sfsnid} introduces a semi-supervised framework with spatial-frequency interaction and realistic brightness constraints, while NightHaze~\cite{nighthaze} uses self-prior learning to improve robustness on real nighttime scenes. Earlier works also emphasized explicit modeling of glow and illumination effects~\cite{li2015nighttime,liu2022vdm}. These studies highlight the importance of degradation-aware modeling for nighttime restoration.

\begin{table*}[!t]
\setlength{\abovecaptionskip}{2mm}
\centering

\resizebox{\textwidth}{!}{%
\begin{tabular}{c|cc|cc|cc|cc}
\hline
\multirow{2}{*}{\textbf{Methods}} 
& \multicolumn{2}{c|}{NH-HAZE~\cite{nhhaze}} 
& \multicolumn{2}{c|}{NH-HAZE2~\cite{nhhaze2}} 
& \multicolumn{2}{c|}{Dense-Haze~\cite{densehaze}} 
& \multicolumn{2}{c}{Challenge-Data} \\
\cline{2-9}
& PSNR$\uparrow$ & SSIM$\uparrow$
& PSNR$\uparrow$ & SSIM$\uparrow$
& PSNR$\uparrow$ & SSIM$\uparrow$
& PSNR$\uparrow$ & SSIM$\uparrow$ \\
\hline

SFSNiD (CVPR 2024)~\cite{cong2024sfsnid}
& 17.318 & 0.523
& 18.017 & 0.720
& 15.201 & 0.392
&---  &---  \\

SFMN (TIP 2025)~\cite{sfmn}
& 20.366 & 0.687
& 18.878 & 0.776
& 16.855 & 0.587
&---  &---  \\

DWT-FFC (CVPR 2023)~\cite{dwtffc}
& 21.621 & \blue{0.706}
& 21.383 & \blue{0.846}
& 17.103 & 0.588
&---  &---  \\

DehazeDCT (CVPR 2024)~\cite{dehazedct}
& \blue{21.632} & 0.702
& \blue{22.345} & \blue{0.846}
& \blue{17.861} & \blue{0.592}
& \blue{27.465} & \blue{0.899} \\

\hline
\textbf{HistoFusionNet (Ours)}
& \textbf{\red{21.972}} & \textbf{\red{0.710}}
& \textbf{\red{22.415}} & \textbf{\red{0.847}}
& \textbf{\red{18.314}} & \textbf{\red{0.611}}
& \textbf{\red{27.879}} & \textbf{\red{0.905}} \\
\hline
\end{tabular}%
}

\caption{Quantitative comparisons between our proposed HistoFusionNet and SOTA methods. Our proposed method achieves superior performance in terms of PSNR and SSIM across four datasets. These numbers are obtained from training with their released code. [Key: \textbf{\red{Best}}, \blue{Second Best}, $\uparrow$ ($\downarrow$): Larger (smaller) values leads to better performance, Challenge-Data: Official dataset for NTIRE 2026 Nighttime Image Dehazing Challenge]}
\label{table_quant_compar}
\vspace{-2mm}
\end{table*}

\Paragraph{Transformer-Based Dehazing}
Transformer-based architectures have shown strong performance in image restoration due to their ability to model long-range dependencies and global context. Models such as Uformer~\cite{Uformer}, SwinIR~\cite{swinir}, AWR-VIP\cite{awr_make}, SG-LLIE~\cite{sgllie}, ReHiT~\cite{rehit}, and Restormer~\cite{restormer} demonstrate the effectiveness of hierarchical and efficient transformer designs across restoration tasks~\cite{liu2021swin}. In dehazing, DeHamer~\cite{guo2022dehamer} combines CNN and transformer representations through transmission-aware 3D position embedding, while DehazeFormer~\cite{dehazeformer} adapts a Swin-style design for haze removal. More recent challenge-oriented methods explore stronger backbones for non-homogeneous dehazing, such as DehazeDCT~\cite{dehazedct}, which uses deformable convolution based transformer-like blocks for adaptive aggregation. Histoformer~\cite{histoformer} further introduces histogram self-attention to group features by intensity, enabling long-range interactions among similarly degraded regions. Such dynamic-range aware modeling is particularly relevant to nighttime scenes, where haze and illumination often induce correlated brightness distortions across distant regions.

\Paragraph{Frequency and Deformable Modeling for Image Restoration}
Frequency-domain modeling has attracted increasing attention in image restoration. Prior works explored frequency-aware restoration through wavelet decomposition, frequency-domain losses, and selective band processing~\cite{freq_yang,freq_jiang,freq1,fsdgn, dong2024shadowrefiner, varlide}. In dehazing, DWT-FFC~\cite{dwtffc} and SFSNiD~\cite{cong2024sfsnid} further show the benefits of spatial-frequency modeling. Meanwhile, deformable convolution enables adaptive spatial sampling and has proven effective for modeling geometric and content-dependent variation~\cite{DCN,DCNv2,DCNv3,xiong2024efficient}, with successful applications to video restoration~\cite{deformable_restore}, super-resolution~\cite{deformable_sr}, exposure correction~\cite{ecmamba}, and dehazing~\cite{aecrnet,dehazedct}. In parallel, AdaIR~\cite{adair} shows that adaptive low- and high-frequency modulation improves restoration across degradations. Inspired by these advances, our method combines DCNv4-based feature extraction, histogram-guided fusion, and frequency-adaptive refinement for nighttime image dehazing~\cite{adair,fsdgn}.

\begin{figure*}[!t]
    \setlength{\abovecaptionskip}{1mm}
    \centering
    \includegraphics[width=0.80\textwidth]{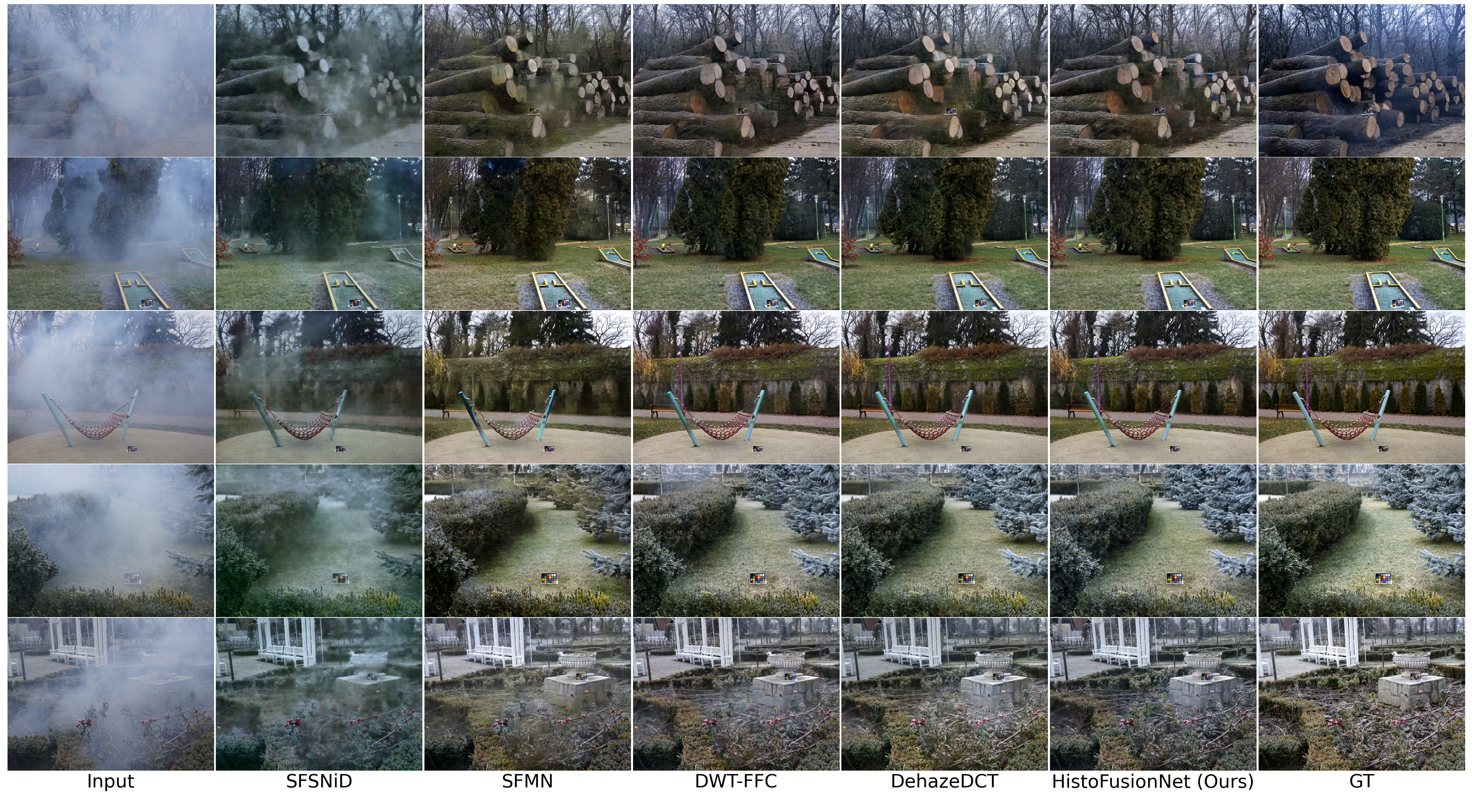}
    \caption{Visual comparisons on NH-HAZE dataset. Compared to other models, our method exhibits higher color fidelity and effective dehazing, yielding compelling results. }
    \label{ntire20}
    \vspace{-2mm}
\end{figure*}

\section{Methods}
\label{sec:method}

The core idea of our method is to combine a strong dehazing network with degradation-aware feature modeling and frequency-adaptive post-refinement. Recent challenge-oriented methods have shown that a two-branch design, jointly exploiting spatial aggregation and complementary frequency cues, is effective for handling complex non-homogeneous degradations~\cite{dehazedct,cong2024sfsnid}. Motivated by the strong efficiency and representation ability of Deformable Convolution v4 (DCNv4)~\cite{xiong2024efficient}, we build our main dehazing backbone on DCNv4-based blocks. As illustrated in Fig.~\ref{fig_frame}, \textbf{HistoFusionNet} adopts a U-shaped dehazing network with a DCNv4-based main branch and an auxiliary frequency-aware branch, followed by a lightweight frequency-adaptive refinement module. The overall optimization is performed in two stages: the dehazing stage first learns the histogram-guided dehazing network (Sec.~\ref{sec_dehazing_module}, Sec.~\ref{sec_loss}), and the refinement stage then fine-tunes the model with the lightweight frequency-adaptive refinement module (Sec.~\ref{sec_refinement_module}).

\subsection{Histogram-Guided Dehazing Network}
\label{sec_dehazing_module}

Given a nighttime hazy input image $\mbf{I}\in\mathbb{R}^{H\times W\times 3}$, we first extract multi-scale features using a U-shaped encoder--decoder architecture, as shown in Fig.~\ref{fig_frame}. The encoder progressively maps the input into latent features $\mbf{F}_{i}\in\mathbb{R}^{\frac{H}{s_i}\times\frac{W}{s_i}\times d_i}$, where $s_i$ and $d_i$ denote the spatial downsampling ratio and feature dimension at the $i$-th scale, respectively. At each encoder stage, several DCNFormer blocks are adopted for representation learning. Similar to the DCNv4-based transformer-like design in~\cite{dehazedct}, each DCNFormer block follows a residual transformer-style structure but replaces conventional self-attention with DCNv4~\cite{xiong2024efficient}, which provides adaptive spatial aggregation with significantly improved efficiency.

For an input feature map $\mbf{X}$ and a reference position $p_0$, the deformable aggregation can be written as
\begin{equation}
\mathbf{y}(p_0)=\sum_{g=1}^{G}\sum_{k=1}^{K}\mathbf{w}_{g}\mathbf{m}_{gk}\mathbf{x}_{g}(p_0+p_k+\Delta p_{gk}),
\label{eq:dcnv4}
\end{equation}
where $G$ is the number of groups and $K$ is the number of sampling points. $\mathbf{w}_g$ denotes the projection weights for the $g$-th group, $\mathbf{m}_{gk}$ is the modulation scalar for the $k$-th sampling point in the $g$-th group, $p_k$ denotes the predefined offset of the $k$-th sampling location relative to the reference position $p_0$, and $\Delta p_{gk}$ is the learned offset. Compared with DCNv3, DCNv4 removes the softmax normalization over modulation weights and improves memory access efficiency, leading to faster convergence and stronger dynamic modeling ability~\cite{xiong2024efficient,DCNv3}. This makes it particularly attractive for nighttime dehazing, where degradations are spatially varying and coupled with local illumination.


To further strengthen global degradation-aware feature interaction, we incorporate \textbf{Histogram Transformer Blocks} at the bottleneck of the U-shaped network. This design is inspired by Histoformer~\cite{histoformer}, which shows that adverse-weather degradations often induce similar brightness attenuation and occlusion patterns across spatially distant regions. Instead of performing attention on fixed local windows or across all tokens indiscriminately, histogram self-attention groups features according to their dynamic-range statistics and performs attention among tokens with similar intensity characteristics.

Concretely, given a bottleneck feature $\mbf{F}\in\mathbb{R}^{h\times w\times c}$, we first compute an intensity descriptor for each spatial token:
\vspace{-1.5mm}
\begin{equation}
s(p)=\frac{1}{c}\sum_{j=1}^{c}\mbf{F}_{p,j},
\label{eq:score}
\end{equation}
where $p$ indexes the spatial position and $j$ indexes the channel dimension. According to the sorted values of $s(p)$, the tokens are rearranged and partitioned into $B$ histogram bins:
\begin{equation}
\{\mbf{F}^{(1)},\mbf{F}^{(2)},\ldots,\mbf{F}^{(B)}\}=\mathcal{P}(\mbf{F}),
\label{eq:partition}
\end{equation}
where $\mathcal{P}(\cdot)$ denotes the sorting and partition operation. For each bin $b$, self-attention is then computed as:
\begin{equation}
\mathrm{Attn}(\mbf{Q}^{(b)},\mbf{K}^{(b)},\mbf{V}^{(b)})
=\mathrm{Softmax}\!\left(\frac{\mbf{Q}^{(b)}(\mbf{K}^{(b)})^{\top}}{\sqrt{d}}\right)\mbf{V}^{(b)},
\label{eq:histo_attn}
\end{equation}
where $\mbf{Q}^{(b)}$, $\mbf{K}^{(b)}$, and $\mbf{V}^{(b)}$ are the query, key, and value embeddings of the $b$-th bin, respectively, and $d$ denotes the embedding dimension. The attended features from all bins are then inversely permuted back to the original spatial order:
\vspace{-1.5mm}
\begin{equation}
\hat{\mbf{F}}=\mathcal{P}^{-1}\!\left(\mathrm{Concat}\big(\hat{\mbf{F}}^{(1)},\ldots,\hat{\mbf{F}}^{(B)}\big)\right).
\label{eq:inverse_partition}
\end{equation}

Each histogram transformer block follows a pre-normalized residual design, consisting of dynamic-range histogram self-attention and a dual-scale gated feed-forward network~\cite{histoformer}. We place these blocks at the bottleneck for efficiency and effectiveness. The reduced spatial resolution lowers the cost of token sorting and attention computation, while the bottleneck features retain rich multi-scale context. This makes the bottleneck well suited for modeling long-range dependencies among similarly degraded regions, which is particularly beneficial in nighttime scenes with haze, glow, and non-uniform illumination. In parallel, we employ an auxiliary frequency-aware branch, inspired by recent dehazing methods that exploit spatial-frequency complementarity~\cite{dehazedct,cong2024sfsnid,dwtffc}. In the dehazing stage, its features are integrated with the decoder through skip connections, providing complementary spatial-frequency guidance for coarse restoration.

\begin{figure*}[t]
    \setlength{\abovecaptionskip}{1mm}
    \centering
    \includegraphics[width=0.80\textwidth]{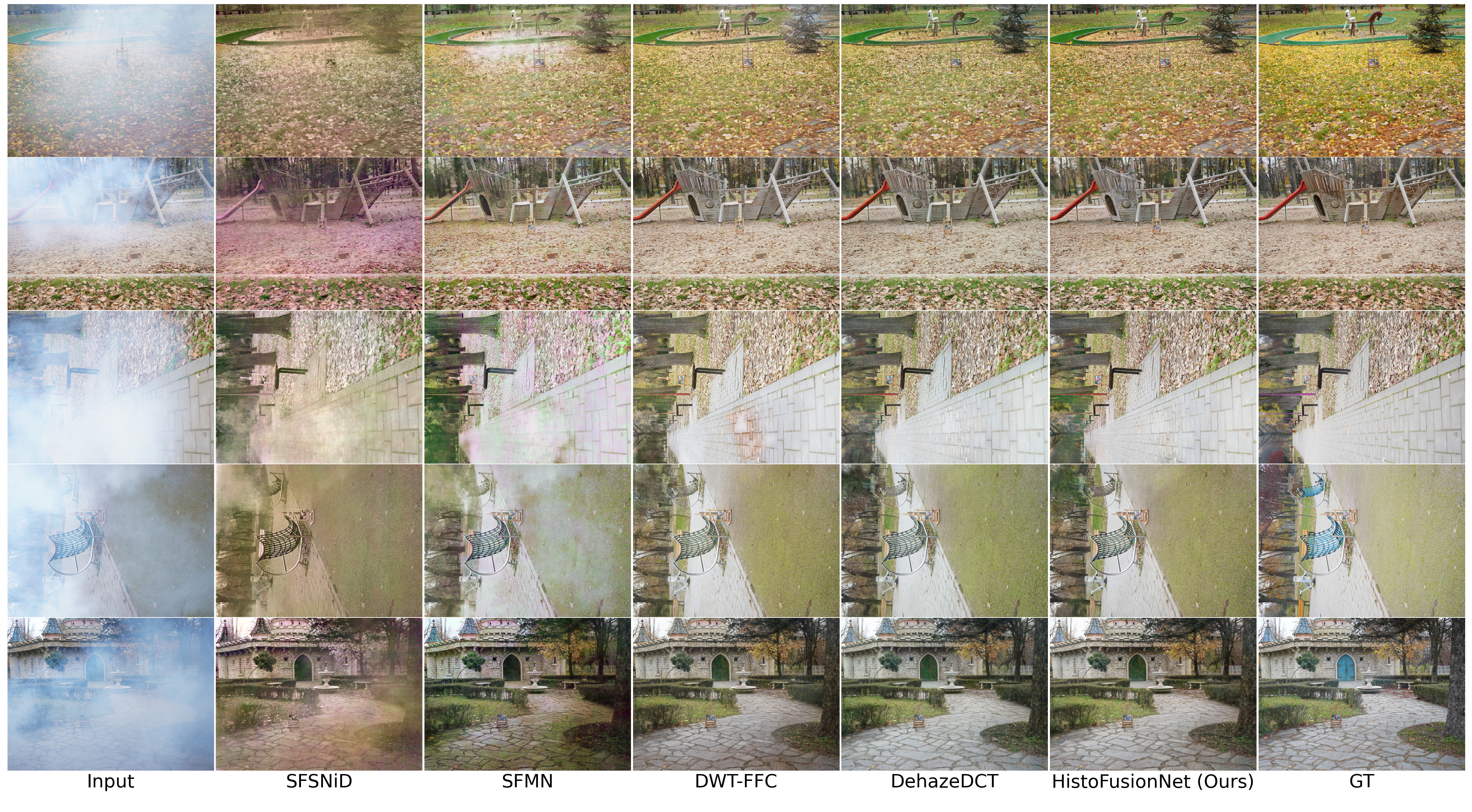}

    \caption{Visual experiment results on NH-HAZE2 dataset. Obviously, our method demonstrates superior performance on color preservation and detail maintaining, further enhancing the overall quality of the output.}
    \label{ntire21}
    \vspace{-2mm}
\end{figure*}

\subsection{Loss Function}
\label{sec_loss}

Following DehazeDCT~\cite{dehazedct}, we optimize the dehazing module with a combination of pixel-wise, structural, perceptual, and adversarial losses:
\begin{equation}
\mathcal{L}_{\mathrm{dehaze}}
=
\mathcal{L}_{1}
+
\alpha \mathcal{L}_{\mathrm{SSIM}}
+
\beta \mathcal{L}_{\mathrm{Percep}}
+
\gamma \mathcal{L}_{\mathrm{adv}},
\label{eq:loss_total}
\end{equation}
where $\mathcal{L}_{1}$, $\mathcal{L}_{\mathrm{SSIM}}$, $\mathcal{L}_{\mathrm{Percep}}$, and $\mathcal{L}_{\mathrm{adv}}$ denote the $L_1$ loss, SSIM loss, perceptual loss, and adversarial loss, respectively. $\alpha$, $\beta$, and $\gamma$ are hyper-parameters, which are set to $0.2$, $0.01$, and $0.0005$ in our implementation.

\subsection{Frequency-Adaptive Refinement}
\label{sec_refinement_module}

Although the dehazing network removes most haze and glow corruption, the restored outputs may still contain residual color deviation, incomplete fine textures, or locally inconsistent enhancement. To further improve the results, we employ a lightweight \emph{Frequency-Adaptive Refinement Module} in the refinement stage on top of the dehazing network, as shown in Fig.~\ref{fig_frame}. This design is motivated by the frequency-adaptive refinement strategy in AdaIR~\cite{adair}, which shows that low- and high-frequency components are affected differently by image degradations and thus provide complementary cues for restoration.

Let $\mbf{F}_{d}=\bm{\theta}(\mbf{I}_{hazy})$ denote the feature produced by the dehazing network $\bm{\theta}(\cdot)$. In the refinement stage, we retain the auxiliary frequency-aware branch and further refine the decoder-side representation through the \emph{M} nodes shown in Fig.~\ref{fig_frame}, which perform adaptive feature fusion at multiple scales. Specifically, for the $i$-th refinement scale, let $\mbf{F}^{i}_{enc}$ and $\mbf{F}^{i}_{dec}$ denote the encoder-transferred feature and the current decoder feature, respectively. Their mixed representation is computed as:
\vspace{-1.5mm}
\begin{equation}
\mbf{F}^{i}_{mix}
=
\alpha_i \mbf{F}^{i}_{enc}
+
(1-\alpha_i)\mbf{F}^{i}_{dec},
\qquad
\alpha_i=\sigma(\theta_i),
\label{eq:mix_refine}
\end{equation}
where $\theta_i$ is a learnable scalar and $\sigma(\cdot)$ denotes the sigmoid function. Inspired by the adaptive mix-up design in GLARE~\cite{glare, gppllie}, this lightweight adaptive mix fusion allows the refinement stage to dynamically balance structural encoder cues and progressively restored decoder features.

\begin{figure*}[!t]
    \setlength{\abovecaptionskip}{1mm}
    \centering
    \includegraphics[width=0.85\textwidth]{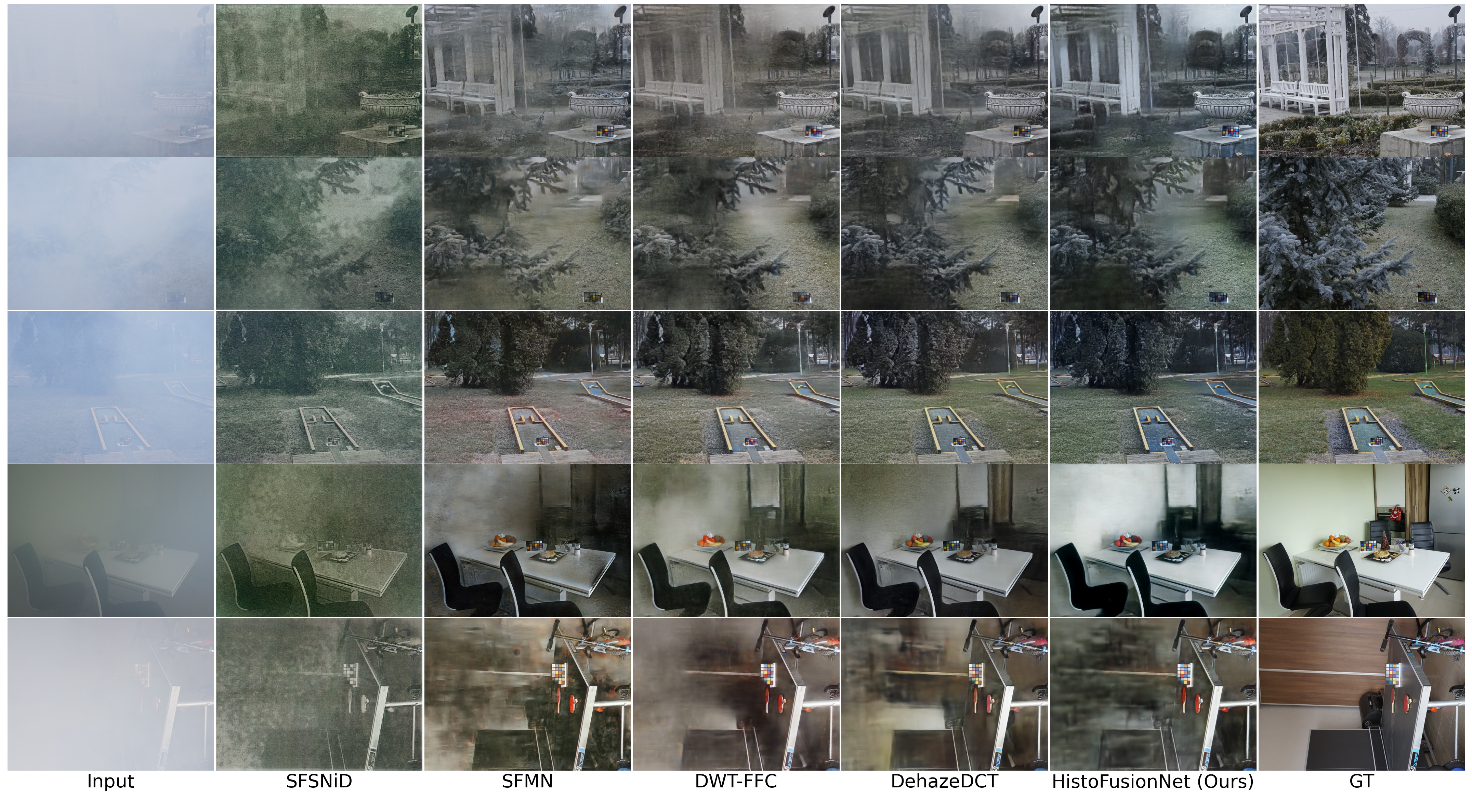}

    \caption{Qualitative comparison on the Dense-Haze dataset. Our method produces clearer structures, better color fidelity, and more faithful detail recovery, leading to higher overall visual quality.}
    \label{ntire23}
    \vspace{-2mm}
\end{figure*}

\begin{table*}[!t]
    \setlength{\abovecaptionskip}{2mm}
    \centering
    \scalebox{0.94}{
    \begin{tabular}{c|cccccc|c}
    \hline
        
    \textbf{Team} & PSNR$\uparrow$ & SSIM$\uparrow$ & LPIPS$\downarrow$ & MUSIQ$\uparrow$ & NIQE$\downarrow$ & FID$\downarrow$ & Final Rank$\downarrow$ \\ 
    \hline
    HistoDeHaze   & \red{\textbf{27.88}} & \red{\textbf{0.91}} & \red{\textbf{0.10}} & 60.40 & 3.77 & \blue{72.25} & \red{\textbf{1}} \\
    XJRes         & \blue{27.44} & \red{\textbf{0.91}} & 0.11 & 63.26 & 3.99 & \red{\textbf{64.73}} & \blue{2} \\
    KETI          & 27.30 & 0.90 & 0.11 & 64.05 & \blue{3.68} & 74.47 & 3 \\
    crivoic       & 26.84 & 0.90 & \blue{0.11} & \blue{64.54} & \red{\textbf{3.64}} & 90.65 & 4 \\
    mi\_camera    & 26.83 & 0.89 & 0.12 & \red{\textbf{65.25}} & 3.95 & 86.06 & 5 \\
    \hline
    \end{tabular}
    }    
\caption{Final ranking (top 5 teams) of the NTIRE 2026 Nighttime Dehazing Challenge. Red and blue indicate the best and second-best results, respectively. Final Rank is reported directly from the official results. [Key: \textbf{\red{Best}}, \blue{Second Best}, $\uparrow (\downarrow)$: larger (smaller) is better.]}
\label{challenge_ranking_2026}
\end{table*}

To provide complementary frequency guidance, the refinement module also decomposes features into low- and high-frequency components in the Fourier domain:
\begin{equation}
\begin{aligned}
\mbf{F}_{low}
&=
\mathcal{F}^{-1}\!\big(\mbf{M}\odot \mathcal{F}(\mbf{F}_{d})\big),\\
\mbf{F}_{high}
&=
\mathcal{F}^{-1}\!\big((1-\mbf{M})\odot \mathcal{F}(\mbf{F}_{d})\big),
\end{aligned}
\label{eq:fft_decomp}
\end{equation}
where $\mathcal{F}(\cdot)$ and $\mathcal{F}^{-1}(\cdot)$ denote the Fourier transform and inverse Fourier transform, $\odot$ denotes element-wise multiplication, and $\mbf{M}$ is an adaptive mask. The low-frequency component mainly preserves global illumination and color information, while the high-frequency component captures edges, textures, and local structures. These complementary cues are then used to refine the restoration features, yielding
\begin{equation}
\hat{\mbf{F}}
=
\mathcal{G}(\mbf{F}_{mix},\mbf{F}_{low},\mbf{F}_{high}),
\label{eq:refine_fuse}
\end{equation}
where $\mathcal{G}(\cdot)$ denotes the frequency-adaptive fusion operator.

The final dehazed image is obtained by
\begin{equation}
\mbf{I}_{out}
=
\bm{\phi}\big(\bm{\theta}(\mbf{I}_{hazy})\big),
\label{eq:final_output}
\end{equation}
where $\bm{\theta}$ and $\bm{\phi}$ denote the dehazing network and the frequency-adaptive refinement module, respectively. Since this stage focuses on residual enhancement rather than full restoration, it introduces only limited overhead while effectively improving local details, structural clarity, and color fidelity in nighttime scenes with complex illumination.

\Section{Experiments}
\label{sec_ex}

\subsection{Experiment Settings}
\Paragraph{Datasets} We quantitatively and qualitatively evaluate our method on four real-world datasets: the NTIRE 2026 Nighttime Image Dehazing Challenge dataset~\cite{ntire2026dehazingdataset}, NH-HAZE~\cite{nhhaze}, NH-HAZE2~\cite{nhhaze2}, and Dense-Haze~\cite{densehaze}. The challenge dataset serves as the main benchmark for our target task, while the other three datasets are used to assess restoration performance and generalization. NH-HAZE and Dense-Haze each contain 55 paired hazy and clean images at resolution $1200\times1600$, where 50 pairs are used for training and 5 for testing. NH-HAZE2 contains 25 paired images of the same resolution, with 20 pairs for training and 5 for testing.



\begin{figure*}[t]
    \setlength{\abovecaptionskip}{2mm}
    \centering
    \includegraphics[width=0.95\textwidth]{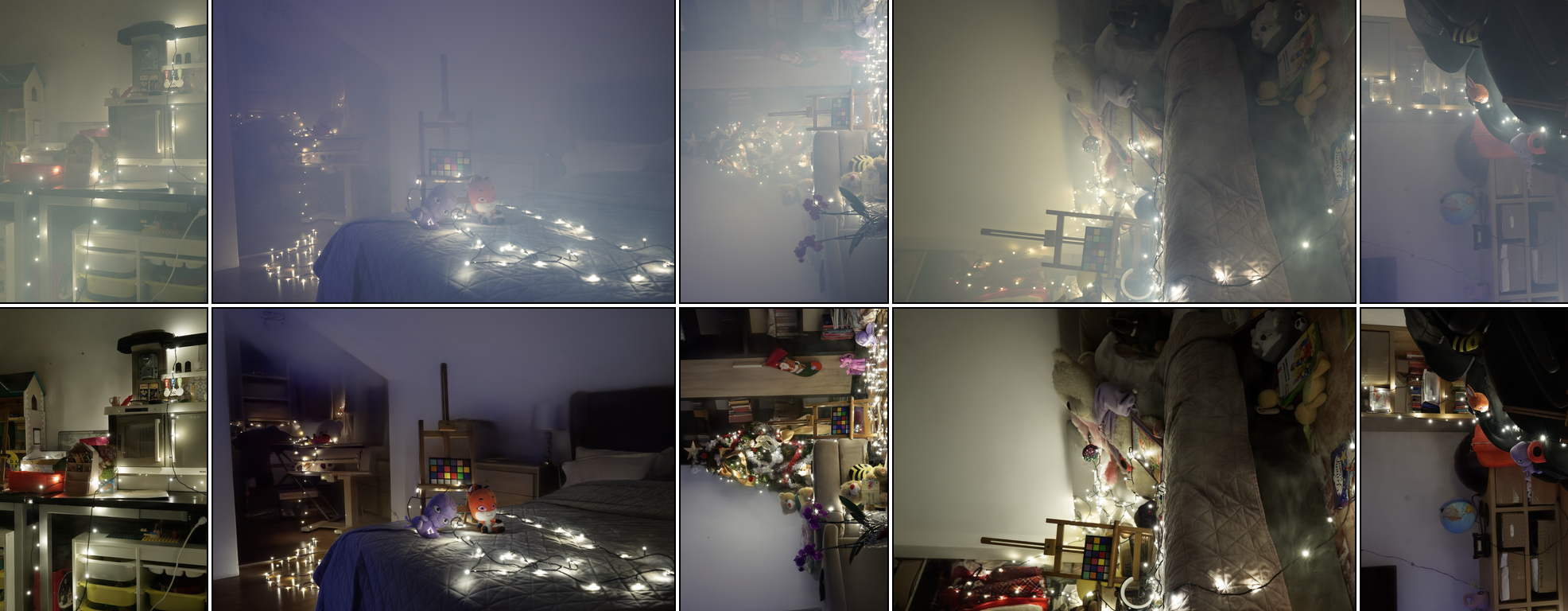}
    \caption{Our results on the validation set of the NTIRE 2026 Nighttime Image Dehazing Challenge, demonstrating strong dehazing performance with improved visibility, clearer structures, and more faithful color restoration.}
    \label{fig:validation_results}
\end{figure*}


\Paragraph{Implementation Details} We implement our method in PyTorch and train it on an NVIDIA H100 80GB HBM3 GPU. During training, paired images are randomly cropped into $384\times384$ patches and augmented by random rotations of 90, 180, or 270 degrees. The optimization process consists of two stages. In stage I, we exclude the refinement module and optimize the dehazing network together with the discriminator using the Adam optimizer with decay factors $\beta_1=0.9$ and $\beta_2=0.999$ for a total of 5,000 epochs. The initial learning rate is set to $1\times10^{-4}$ and is decayed by a factor of 0.5 at 2000, 3000, and 4000 epochs. In stage II, we further optimize the refinement module for 200 epochs using a fixed learning rate of $1\times10^{-5}$.

\subsection{Comparisons with SOTA Methods}

\Paragraph{Compared Methods and Evaluation Metrics}
We compare \textbf{HistoFusionNet} with four recent dehazing baselines, including SFSNiD~\cite{cong2024sfsnid}, SFMN~\cite{sfmn}, DWT-FFC~\cite{dwtffc}, and DehazeDCT~\cite{dehazedct}. For quantitative evaluation, we use two full-reference metrics, namely Peak Signal-to-Noise Ratio (PSNR) and Structural Similarity Index Measure (SSIM), to assess reconstruction quality and structural fidelity.

\Paragraph{Quantitative Comparisons}
Tab.~\ref{table_quant_compar} reports the quantitative comparisons on four datasets. As shown, \textbf{HistoFusionNet} achieves the best overall performance in terms of both PSNR and SSIM across all benchmarks. Specifically, it achieves 27.879 dB PSNR and 0.905 SSIM on the NTIRE 2026 Nighttime Image Dehazing Challenge dataset. Compared with the second-best method, our model improves PSNR by 0.340 dB, 0.070 dB, 0.453 dB, and 0.414 dB on the four datasets, respectively, while achieving higher or competitive SSIM in all cases.

\Paragraph{Qualitative Comparisons}
The visual comparisons in Figs.~\ref{ntire20},~\ref{ntire21}, and~\ref{ntire23} further demonstrate the advantage of \textbf{HistoFusionNet}. Compared with other methods, our approach produces clearer structures, more faithful colors, and fewer residual haze artifacts. In challenging cases, competing methods often leave haze residue, introduce color shifts, or over-smooth fine details, whereas \textbf{HistoFusionNet} yields more natural and visually consistent results.

\begin{table}[!t]
    \setlength{\abovecaptionskip}{2mm}
    \centering
    \scalebox{1}{
    \begin{tabular}{c|cc}
    \hline
        
    \textbf{Configurations} &PSNR$\uparrow$  &SSIM$\uparrow$ \\ 
    \hline
    w/o frequency-adaptive refinement &18.103 &0.600 \\    
    w/o histogram transformer blocks &17.861 &0.592 \\
    w/o frequency-aware branch &17.984 &0.597 \\
    \hline
    \end{tabular}
    }    
\caption{The ablation result of our method on the Dense-Haze dataset. Each component of our HistoFusionNet contribute positively to our final dehazing performance.}
\label{tab_abla}
\end{table}

\subsection{Ablation Study}
\label{sec_ex_abla}

To analyze the effectiveness of the key components of \textbf{HistoFusionNet}, we conduct ablation experiments on the Dense-Haze dataset. In particular, we focus on the three major modules added in \textbf{HistoFusionNet}, namely the histogram transformer blocks, the frequency-aware branch, and the frequency-adaptive refinement. Specifically, we remove one module at a time while keeping the remaining architecture unchanged, and report the results in Tab.~\ref{tab_abla}.

\Paragraph{Effectiveness of histogram transformer blocks}
To evaluate the contribution of the histogram transformer blocks, we remove them from the bottleneck while keeping the rest of the architecture unchanged. As shown in Tab.~\ref{tab_abla}, this variant achieves 17.861 dB PSNR and 0.592 SSIM on Dense-Haze, both lower than the corresponding results of the full \textbf{HistoFusionNet}. This drop confirms the importance of histogram transformer blocks. By enabling dynamic-range aware global feature interaction, they capture long-range dependencies among similarly degraded regions and improve both reconstruction accuracy and structural fidelity.

\Paragraph{Effectiveness of frequency-adaptive refinement}
To study the effect of the refinement module, we remove it and directly use the output of the dehazing network as the final result. As reported in Tab.~\ref{tab_abla}, this variant obtains 18.103 dB PSNR and 0.600 SSIM, which is still inferior to the full model by 0.211 dB and 0.011 SSIM. This shows that the refinement stage provides complementary enhancement, further improving residual details and color consistency.

\Paragraph{Effectiveness of frequency-aware branch}
To evaluate the effect of frequency-aware branch, we remove it while keeping the histogram transformer blocks and excluding the refinement module. As shown in Tab.~\ref{tab_abla}, this variant achieves 17.984 dB PSNR and 0.597 SSIM, which is inferior to the corresponding dehazing network with the frequency-aware branch. This result verifies that the frequency-aware branch provides beneficial complementary guidance to the main branch.

\subsection{Performance of Our Method on NTIRE 2026 Nighttime Image Dehazing Challenge}

The official challenge results are evaluated using PSNR, SSIM, LPIPS, MUSIQ, NIQE, and FID. Tab.~\ref{challenge_ranking_2026} summarizes the quantitative results of the top 5 teams. Our solution ranks \textbf{1st} overall among 22 participating teams, achieving the best PSNR (27.88 dB), SSIM (0.91), and LPIPS (0.10), together with competitive perceptual quality on the other evaluation metrics. The visual results on the official test and validation sets in Fig.~\ref{fig:figure_challenge} and Fig.~\ref{fig:validation_results} further confirm that our method produces clear, natural, and visually faithful dehazing results under challenging nighttime conditions.

\section{Conclusion}
\label{sec:conclu}
In this paper, we presented \textbf{HistoFusionNet}, an effective network for nighttime image dehazing that combines a DCNv4-based dehazing backbone, histogram-guided fusion, and frequency-adaptive refinement within a unified framework. Specifically, we adopted a U-shaped dehazing architecture with a DCNv4-based main branch and an auxiliary frequency-aware branch to handle the complex spatially varying degradations in nighttime hazy scenes. In addition, histogram transformer blocks were introduced at the bottleneck to model long-range dependencies among regions with similar dynamic-range degradation, while a lightweight frequency-adaptive refinement module was employed to further improve color fidelity and recover fine details. Extensive experiments demonstrate the effectiveness of the proposed method against recent strong baselines and across multiple real-world hazy datasets. Furthermore, \textbf{HistoFusionNet} achieved \textbf{1st place} in the NTIRE 2026 Nighttime Image Dehazing Challenge among 22 participating teams.

\clearpage
{
    \small
    \bibliographystyle{ieeenat_fullname}
    \bibliography{main3}
}
\end{document}